\documentclass{article}
\usepackage[usenames,dvipsnames,svgnames]{xcolor}
\usepackage{hyperref}
\hypersetup{
    colorlinks=true,
    citecolor=MidnightBlue,
    urlcolor=Bittersweet,
}
\usepackage[margin=1in]{geometry}
\usepackage{amsmath,amssymb}
\usepackage{graphicx}
\usepackage{bm} 
\usepackage{subcaption} 
\usepackage{tikz}
\usetikzlibrary{spy}
\usepackage{ulem}
\usepackage{textcomp}
\DeclareMathOperator*{\argmin}{arg\,min}

\newcommand{\x}{\mathbf{x}}
\newcommand{\z}{\mathbf{z}}
\newcommand{\X}{\mathbf{X}}	
\newcommand{\y}{\mathbf{y}}
\newcommand{\A}{\mathbf{A}}
\renewcommand{\z}{\mathbf{z}}
\newcommand{\W}{\mathbf{W}}

\newcommand{\omg}{\mathbf{\Omega}}

\newcommand{\Z}{\mathbf{Z}}
\newcommand{\eps}{\boldsymbol{\varepsilon}}

\newcommand{\R}{\mathbb{R}}

\renewcommand{\vec}[1]{\mathbf{#1}}
\newcommand{\mat}[1]{\mathbf{#1}}

\newcommand*\samethanks[1][\value{footnote}]{\footnotemark[#1]}
\begin{document}

\title{Model-based Reconstruction with Learning:  From Unsupervised to Supervised and Beyond}
\author{Zhishen Huang\thanks{Department of Computational Mathematics, Michigan State University, East Lansing, MI 48824, USA, \newline\{huangz78, mccann13, ravisha3\}@msu.edu}\and Siqi Ye\thanks{University of Michigan - Shanghai Jiao Tong University Joint Institute, Shanghai Jiao Tong University,
Shanghai 200240, China, yesiqi@sjtu.edu.cn} \and Michael T. McCann\samethanks[1] \and Saiprasad Ravishankar\samethanks[1]}

\maketitle

\begin{abstract}
    Many techniques have been proposed for image reconstruction in medical imaging that aim to recover high-quality images especially from limited or corrupted measurements. Model-based reconstruction methods have been particularly popular (e.g., in magnetic resonance imaging and tomographic modalities) and exploit models of the imaging system's physics together with statistical models of measurements, noise and often relatively simple object priors or regularizers. For example, sparsity or low-rankness based regularizers have been widely used for image reconstruction from limited data such as in compressed sensing. Learning-based approaches for image reconstruction have garnered much attention in recent years and have shown promise across biomedical imaging applications. These methods include synthesis dictionary learning, sparsifying transform learning, and different forms of deep learning involving complex neural networks. We briefly discuss classical model-based reconstruction methods and then review reconstruction methods at the intersection of model-based and learning-based paradigms in detail. This review includes many recent methods based on unsupervised learning, and supervised learning, as well as a framework to combine multiple types of learned models together.
\end{abstract}

\label{chap1}

\section{Introduction}
Imaging modalities such as magnetic resonance imaging (MRI), X-ray computed tomography (CT),
positron-emission tomography (PET), and single-photon emission computed tomography (SPECT) are regularly used in clinical practice for characterizing anatomical structures and physiological functions and to aid clinical diagnosis and treatment.
Various sophisticated image reconstruction methods have been proposed for these modalities that help recover high-quality images especially from limited or corrupted measurements.
Classical analytical reconstruction methods include filtered back-projection (FBP) for CT and the inverse Fourier transform for MRI.
These methods rely on simple imaging models and have efficient implementations,
but can product suboptimal reconstructions,
especially when the number of measurements is limited.

Model-based image reconstruction (MBIR, alternatively called model-based iterative reconstruction) methods are popular in many medical imaging modalities. These methods exploit models of the imaging system's physics (forward models) along with statistical models of the measurements and noise and often simple object priors.
They iteratively optimize model-based cost functions to estimate the underlying unknown image~\cite{sauer:93:alu,thibault:06:arf}. Typically, such cost functions consist of a data-fidelity term (e.g., least squares or weighted least squares penalties) capturing the imaging forward model and the measurement/noise statistical model and a regularizer term (e.g., smoothness or sparsity penalties) capturing assumed object properties.
For example, sparsity or low-rankness based regularizers have been widely used for image reconstruction from limited data (e.g., in compressed sensing~\cite{donoho2006compressed}).

Data-driven and learning-based approaches have gained much interest in recent years for biomedical image reconstruction. These methods learn representations of biomedical images and use them in MBIR or learn complex mappings between limited or corrupted measurements and high-quality images.
These include methods such as synthesis dictionary learning~\cite{sai2011dlmri,xu:12:ldx} and sparsifying transform learning~\cite{sai2013tl,wensailukebres19} and the growing domain of deep learning~\cite{schlemper18,saireview20}.
In particular, there has been increasing interest in methods that leverage both learning-based and MBIR tools.
This chapter reviews such methods in detail.

The chapter is organized as follows. Section~\ref{classicalmbir} reviews classical MBIR methods. Sections~\ref{mbirunsupervised} and~\ref{mbirsupervised} review methods combining MBIR with unsupervised and supervised learning, respectively, while Section~\ref{mbircombined} reviews methods that combine MBIR with multiple paradigms or ways of machine learning. Finally, in Section~\ref{discussion}, we summarize our perspectives on these methods and future directions.

\section{Classical Model-based Image Reconstruction} \label{classicalmbir}
We start the discussion of model-based iterative reconstruction (MBIR) with the mathematical formulation of the image reconstruction problem. 
An image is represented as an array of discrete pixels and denoted as a vector $\x$ of length $mn$ (the number of pixels, where $m$ and $n$ are the number of rows and columns in the image), which can assume either real or complex values depending on the imaging modality. Classical imaging systems involve a measurement or forward operator $\mathcal{A}$ applied to an image for reconstruction, and one obtains corresponding observations $\y = \mathcal{A}(\x) + \eps$, where $\eps$ is the noise in the measurements. When the measurement process is linear, the application of the sensing or measurement operator on an image can often be characterized by a matrix-vector product. Early publications on MBIR methods tended to exploit the mathematical Bayesian framework for image reconstruction. The dominant Bayesian approach~\cite{Hanson_Bayes93} for reconstructing an image $\x$ from measurements $\y$ is the maximum a posteriori (MAP) estimation, which finds the maximizer of the posterior distribution $p(\x|\y)$ by Bayes law, i.e.
\begin{equation}
\label{prob::Bayes_reconstructor}
\widehat{\x} = \mathrm{arg}\min_\x\, -\log p(\y|\x) - \log p(\x) 
\end{equation}
where $-\log p(\y|\x)$ is the negative log-likelihood that encompasses the physics of the imaging system and noise statistics, and $p(\x)$ is the prior which carries the assumptions one makes about the properties of the image under consideration. For any prior $p(\x)$, one can define $\beta \mathcal{R}(\x) = -\log p(\x)$. Thus, the form of the Bayes reconstructor~\eqref{prob::Bayes_reconstructor} can be cast as the solution to the following regularized optimization problem:
\begin{equation}
\label{prob::forward_model}
\min_{\x}\, f(\x,\y) + \beta \mathcal{R}(\x),
\end{equation}
where $f(\x,\y)$ characterizes the data fidelity (e.g.\ $\frac{1}{2}\|\mathcal{A}(\x) - \y\|_2^2$) and $\mathcal{R}(\x)$ is a regularizer that guides the reconstructed image $\x$ to have some assumed properties such as smoothness, sparsity in the frequency domain or wavelet domain, or low-rankness.

\textit{Regularization} provides a mechanism of forcing the reconstructed images to have certain desired properties. When the measurement matrix in the least-square data-fidelity term has a large condition number, it makes the unregularized optimization problem ill-conditioned. Or when it is underdetermined, the corresponding unregularized optimization problem has infinitely many reconstruction results. 
In such cases, the Tikhonov regularizer $\mathcal{R}(\x) = \|\x\|_2^2$ with properly chosen magnitude will render the optimization problem well-posed.
Recent methods use edge-preserving regularization involving non-quadratic functions of the differences between neighboring pixels~\cite{Ahn_differencePen15},  implicitly assuming that the image gradients are sparse.
In 2D, an example of a difference-based regularizer 
would be $\mathcal{R}(\x) = \sum_{l=2}^m\psi\big(\x_{l,:}-\x_{l-1,:}\big) + \sum_{l=2}^n\psi\big(\x_{:,l} - \x_{:,l-1}\big)$, where $\x_{l,:}$ and $\x_{:,l}$ denote the $l$th row and column of the 2D image respectively, and $\psi$ is applied to vector elements (scalars) and summed, e.g., the hyperbola $\psi(z)=\sqrt{|z|^2+\delta^2}$ or a generalized Gaussian function~\cite{Bouman_gaussianPen93}. In practice, twists to the form of the regularizer have been proposed to address the issue of non-uniform spatial resolution in tomography (by introducing spatial weights in the penalty for neighboring pixels~\cite{Fressler_spatialPen96}) and the issue of sensitivity to the choice of hyper-parameters (by penalizing relative difference instead of absolute difference between pixels~\cite{Nuyts_relativeDiffPen02}). These difference-based regularizers are relatives of the well-studied total variation (TV) regularizers, which use the nonsmooth absolute value potential function $\psi(z) = |z|$. TV regularizers impose a strong assumption of gradient sparsity and may be less suitable for piecewise smooth images. In particular, the TV regularizer leads to CT images with undesirable patchy textures.

Another important aspect of classic image reconstruction techniques is the underlying \textit{sampling} that constitutes the measurement, as it is natural to ask what the minimum amount of required measurements is to recover an image with prior assumptions in a given context. Conventional approaches to sampling signals follow Shannon’s celebrated theorem: the sampling rate must be at least twice the maximum frequency present in the signal.
Many natural signals (and in particular images) have concise representations when expressed in a convenient basis, or in other words, the information of a signal concentrates on a few basis vectors in a domain albeit not necessarily bandlimited.
When an image can be \textit{sparsely} represented in a certain domain and the measurements of the image are \textit{incoherent} to the basis of its representation, such measurements may suffice to capture most of the information contained in that image and thus provide an accurate reconstruction of the image through convex optimization. This observation holds notwithstanding the fact that the sampling frequency may be lower than the Nyquist frequency~\cite{CandesRombergTao_CS06}. To formulate this observation, consider a wide rectangular-shaped measurement matrix $\A\in\R^{d\times mn}$ (e.g., in CT) or $\A\in\mathbb{C}^{d\times mn}$ (such as in MRI), where typically $d<mn$. The image reconstruction can be mathematically formulated as 
\begin{equation}
    \label{prob::analysis_regularization}
    \widehat{\x} = \mathrm{arg} \min_{\x} \frac{1}{2}\|\A\x-\y\|^2_2 + \beta \mathcal{R}(\mathbf{W}\x),
\end{equation}
where $\mathbf{W}$ is a sparsifying transform matrix that transforms the input image $\x$ to a domain where it is sparse. 

The $\ell_1$ norm is often used as the sparsity regularizer, and it can be viewed as a convex relaxation or convex envelope of the nonconvex $\ell_0$ `norm' that counts the amount of non-zero elements in a vector. Empirical observation reveals that images often have a sparse representation under Fourier or wavelet transforms, and the \textit{compressed sensing} theory shows that random sensing matrices such as Gaussian matrices or Rademacher matrices are largely incoherent with fixed bases \cite{Foucart_CSBook13}. 
A signature result in compressed sensing theory states that with only $\mathcal{O}(S\log(N/S))$ measurements encoded in a matrix $\mathbf{A}$ with the incoherence property, the convex \textit{basis pursuit} problem $\min_{\widetilde{\x}\in\R^N} \|\widetilde{\x}\|_1$ subject to $\y=\mathbf{A}\widetilde{\x}$ can recover any $S$-sparse signal vector $\x\in\R^N$ \textit{exactly} with high probability.

Low-rank models have also shown promise in imaging applications. The assumption of low-rank structure is particularly useful when processing dynamic or time-series data, and has been popular in dynamic MRI, where the underlying image sequences are correlated over time. Recent works have applied low-rank plus sparse (L+S) model to dynamic MRI reconstruction~\cite{TDAA_dynamicMRI14,Otazo_LowRankSparseMRI15}, and accurate reconstruction is reported when the underlying L and S components are distinguishable and the measurement is also sufficiently incoherent with these components. Low-rank components capture the background or slowly changing parts of the dynamic object, while the sparse components capture the dynamics in the foreground such as local motion or contrast changes. The L+S reconstruction problem is formulated as 
\begin{equation}
\label{prob::LplusS}
    \min_{\x_\text{L}, \x_\text{S}} \frac{1}{2}\|\A(\x_\text{L}+ \x_\text{S}) - \y\|_2^2 + \lambda_\text{L} \|\mathrm{Mat}(\x_\text{L})\|_* + \lambda_\text{S} \|\mathbf{W}\x_\text{S}\|_1,
\end{equation}
where $\|\cdot\|_*$ is the nuclear norm, $\x_\text{L}$ and $\x_\text{S}$ are vectors of dimension $mnt$ with $t$ being the number of temporal frames, $\mathrm{Mat}(\x_\text{L})$ reshapes $\x_\text{L}$ into a space-time matrix of dimension $mn\times t$, and $\mathbf{W}$ is a sparsifying operator. 

Another low-rankness based image reconstruction approach proposed for the medical imaging setting uses a low-rank Hankel structure matrix by exploiting the duality between spatial domain sparsity and the spectral domain Hankel matrix rank. The corresponding optimization problem is purely in the measurement domain. Given sparsely sampled spectral measurements on the index set $\Omega\subseteq\{0,1,\cdots,n-2,n-1\}$, the missing spectrum estimation problem is cast as
\begin{align}
\label{prob::lowrankH}
    &\mathrm{arg}\min_{\mathbf{m}\in\mathbb{C}^n} \|H(n,d) (\mathbf{m})\|_*\\
    &\textrm{s.t. } P_{\Omega}(\mathbf{m}) = P_{\Omega} (\widehat{\x}), \nonumber
\end{align}
where  $H(n,d) (\mathbf{m}) =\begin{pmatrix}
m(0) & m(1) & \cdots & m(d-1)\\
\vdots & \vdots & \vdots & \vdots \\
m(n-d) & m(n-d+1) & \cdots & m(n-1)\end{pmatrix} \in\mathbb{C}^{n\times d}$ ($d$ is a hyperparameter), $P_\Omega(\cdot)$ denotes the projection on the measured $k$-space samples on the index set $\Omega$,  and $\widehat{\x}$ is the Fourier data of the signal. When $|\Omega| = \mathcal{O}\big(\textrm{rank}(H(n,d))d^{-1}\log n\big)$, the optimization problem~\eqref{prob::lowrankH} recovers the image exactly with high probability~\cite{Jin_lowrank_Hankel_16}.

Iterative methods such as proximal gradient descent~\cite{Huang_PPD} and the alternating direction method of multipliers (ADMM)~\cite{BeckBook_FirstOrder} are deployed for solving the aforementioned optimization problems. To minimize loss functions with nuclear norm, a surrogate relaxation and careful initialization is needed to deploy the conditional gradient method~\cite{Yu_nuclearnormMin14}. 

For classical MBIR models involving sparsity, the transforms such as the Fourier transform and wavelet transform are fixed and the corresponding measurement operators are generic and nonadaptive. Also, the choice of hyperparameters can largely affect the quality of the reconstructed images while there is not always a systematic guideline for how to choose hyperparameters for imaging problems. In the following sections, object-adaptive MBIR methods will be highlighted.

\begin{figure}[!htp]
    \centering
    \includegraphics[width=1\linewidth]{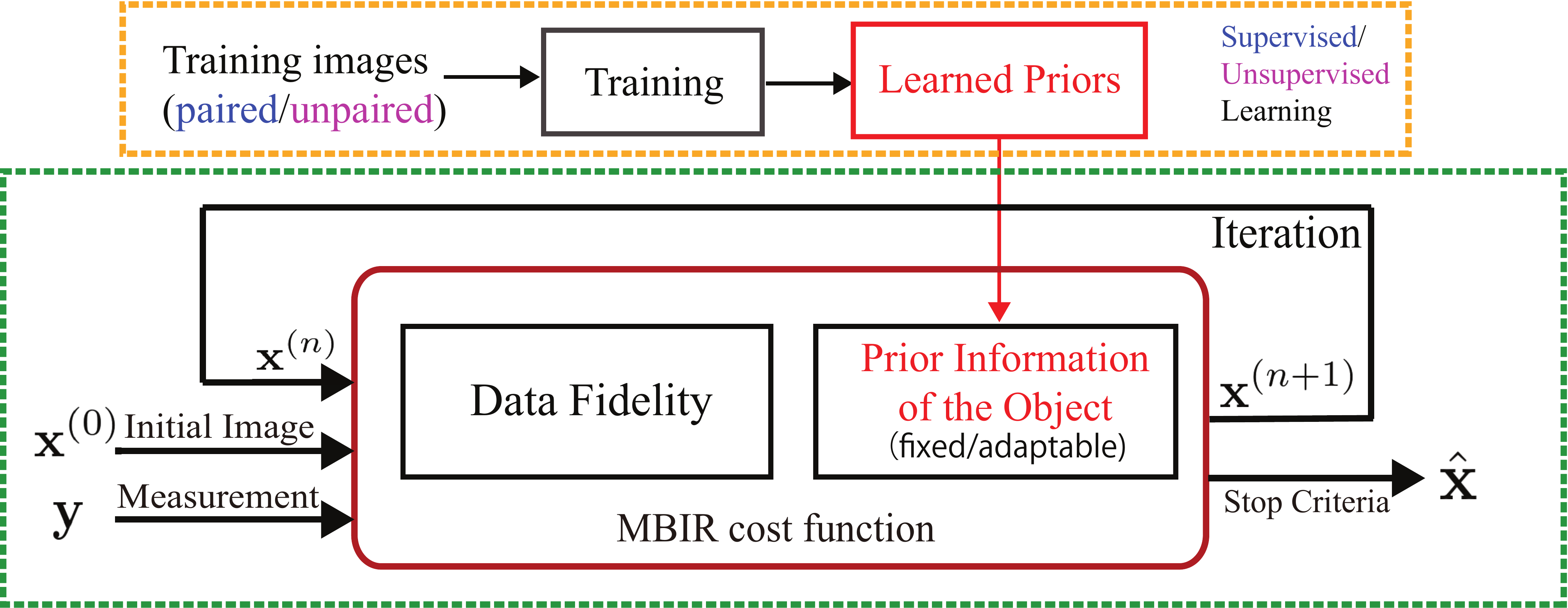}
    \caption{A general illustration of MBIR with learning.}
    \label{fig:MBIR-Learn}
\end{figure}
\section{MBIR with Unsupervised Learning}\label{mbirunsupervised}
In this section, we introduce several unsupervised learning techniques that can be involved in MBIR to provide unsupervised prior knowledge of the target image.
We first introduce two sparsity and compressed sensing-inspired unsupervised learning techniques: dictionary learning and sparsifying transform learning, which can be incorporated into MBIR; then we present some generic neural network-based unsupervised learning techniques and their combinations with MBIR. 
The descriptions are based on applications with real-valued images such as in CT, but they are readily extended to applications with complex-valued images such as MRI.

\subsection{Synthesis Dictionary Learning}
A synthesis dictionary provides a set of 
atoms, sparse linear combinations of which can  represent a signal or an image.
Let $\bm{D}\in\mathbb{R}^{m\times n}$ denote a dictionary, whose columns are atoms.
Usually, the dictionary is overcomplete, i.e., ${n \gg m}$.
A signal $\x \in \mathbb{R}^{m}$ can be represented as ${\x = \bm{D} \z}+\mathbf{e}$, where 
$\z \in\mathbb{R}^{n}$ denotes sparse coefficients, and
$\mathbf{e}\in \mathbb{R}^{m}$ denotes small differences between the synthesized signal and the original one.

Given a matrix $\X \in \mathbb{R}^{m\times S}$ composed of $S$ vectorized training image patches, the dictionary learning problem can be formulated as
\begin{equation}
\textrm{arg}\min_{\bm{D} \in \mathcal{D},\Z} \sum_{i=1}^{S} \big(\|\X_i - \bm{D}\Z_i\|_2^2 + \lambda_i\|\Z_i\|_0\big),
\label{dlproblem}
\end{equation}
where $\mathcal{D}$ denotes the set of feasible dictionaries (e.g., those with unit norm columns~\cite{k-svd:06}), $\X_i$ and $\Z_i$ denote the $i$th columns of matrices $\X$ and $\Z$ respectively, and $\lambda_i$ is a nonnegative regularizer parameter for the $i$th patch. The operator $\|\cdot\|_0$ is the $\ell_0$ sparsity functional that counts the number of nonzeros in a vector.
Sparsity constraints (rather than penalty) are also often used~\cite{k-svd:06,sai2011dlmri}.

The dictionary learning problem (\eqref{dlproblem} or its variants) is often optimized via alternating between a sparse coding step, i.e., updating $\Z$ with fixed dictionary $\bm{D}$, and a dictionary learning step, i.e., updating $\bm{D}$ with fixed sparse coefficients $\Z$. K-SVD~\cite{k-svd:06} is a well-known approximate algorithm for dictionary learning, but it involves computationally expensive sparse coding using the greedy orthogonal matching pursuit~\cite{OMP04} algorithm.
Recent works efficiently tackle dictionary learning with sparsity penalties (e.g., the sum of outer products dictionary learning method~\cite{SOUP-DIL}) using block coordinate descent optimization with thresholding-based closed-form sparse code updates. However,~\cite{SOUP-DIL} still optimizes rows of $\Z$ in a sequential manner, creating some bottleneck.


The image reconstruction problem incorporating a pre-learned dictionary-based regularizer can be formulated as
\begin{equation}
\min_{\x,\z} f(\x,\y) + \beta\sum_{j=1}^{N}\big(\|\mathbf{P}_j\x - \bm{D}\z_j\|_2^2 + \gamma_j\|\z_j\|_0\big),
\label{dictlrecon}
\end{equation}
where $f(\x,\y)$ represents the data-fidelity term (e.g., $f(\x,\y) \triangleq \left \| \A\x - \y \right \|_{2}^{2}$) capturing the imaging measurement model and noise model, $\A$ and $\y$ are the system matrix and measurements respectively as described in Section~\ref{classicalmbir}, $\mathbf{P}_j$ is the patch extraction operator (the dictionary acts on image patches), $\beta>0$ is the regularizer parameter, and the nonnegative $\gamma_j$'s control sparsity levels. While using pre-learned (from a set of images) dictionaries have been shown to be useful for biomedical image reconstruction~\cite{xu:12:ldx}, the dictionary can also be directly learned or optimized at image reconstruction time~\cite{sai2011dlmri,xu:12:ldx,wensailukebres19}, which can yield models highly adaptive to the underlying image. For example,~\eqref{dictlrecon} can be optimized by alternating between updating $\x$ (by solving a least squares problem) and the sparse codes. Dictionary learning methods provided one of the earliest breakthroughs for machine learning in biomedical image reconstruction~\cite{sai2011dlmri,xu:12:ldx,lingaljacob13}.

\subsection{Sparsifying transform learning}	\label{sec:sparsifying-transform}
The sparsifying transform is a general analysis operator that is applied to a signal to produce sparse or approximately sparse representations.
For a sparsifying transform $\omg\in\mathbb{R}^{m\times n}$, the model is
\begin{equation}
\omg \x = \z + \mathbf{e},\quad \|\z\|_0 \leq s,
\end{equation}
where $\x \in \mathbb{R}^{n}$ denotes the original signal, $\z \in \mathbb{R}^{m}$ is a vector with several zeros (or up to $s$ nonzeros), and $\mathbf{e}\in \mathbb{R}^{m}$ represents the approximation error or residual in the transform domain.
An advantage of the sparsifying transform model over the synthesis dictionary is that the signal's sparse coefficients can be cheaply obtained by thresholding~\cite{sai2013tl}.
This has led to very efficient sparsifying transform learning methods 
for image recovery tasks, including image denoising and image reconstruction~\cite{sai2013tl,pfister2014model,pwls-ultra2018,wensailukebres19}.

In the following, we first introduce several types of sparsifying transforms and their learning, and then describe their application to biomedical image reconstruction. 
Sparsifying transforms can be square ($m=n$), overcomplete ($m>n$), or undercomplete ($m<n$). They could also be more general nonlinear operators.
Square sparsifying transforms and their extensions have been adopted in several works. Hence, first, we detail their learning.

\paragraph{Learning Square Transforms}
Given a training matrix $\X \in \mathbb{R}^{n\times S}$, whose columns are the $S$ vectorized patches of length $n$ drawn from training images, the learning problem is formulated as follows:
\begin{equation} \label{eq:ST-learn}
\begin{aligned}
\textrm{arg}\min_{\omg,\Z}\| \omg\mathbf{X}-\Z \|_{F}^2 + \lambda \widetilde{\mathcal{R}}(\omg)+ \sum_{i = 1}^M\gamma^2\|\Z_i\|_0 ,
\end{aligned}
\end{equation}
where $\omg\in \mathbb{R}^{n\times n}$ is the transform to be learned, $\Z \in \mathbb{R}^{n\times S}$ is the transform domain sparse coefficients matrix associated with $\omg$, 
$\lambda>0$ is a weighting parameter for regularizer $\widetilde{\mathcal{R}}(\omg)$, and $\gamma>0$ controls the sparsity of $\Z$.
The regularizer $\widetilde{\mathcal{R}}(\omg)$ is designed to avoid trivial solutions of $\omg$ such as a zero matrix or $\omg$ with repeated rows or arbitrarily large scaling of $\omg$~\cite{sai2013tl}. A possible choice is ${\widetilde{\mathcal{R}}(\omg) \triangleq \| \omg \|_{F}^2 - \log|\det \omg|}$~\cite{sai2013tl}. One could also enforce $\omg$ to be unitary, i.e., $\omg^T\omg = \bm{I}$~\cite{wensailukebres19}. 
Alternating minimization strategies are commonly used to optimize~\eqref{eq:ST-learn}. 
Specifically, when solving for $\Z$ with fixed $\omg$, we have the subproblem ${\argmin_{\Z}\| \omg\mathbf{X}-\Z \|_{F}^2 + \sum_{i = 1}^M\gamma^2\|\Z_i\|_0}$, which can be exactly solved via hard-thresholding, i.e., ${\widehat{\Z}=H_{\gamma}(\omg\X)}$, with $H_{\gamma}(\cdot)$ being an element-wise hard-thresholding operator that zeros out elements smaller than $\gamma$.
When solving for $\omg$ (fixed $\Z$), the corresponding subproblem can also have a closed-form solution~\cite{STlearning15}. These algorithms have proven convergence guarantees~\cite{STlearning15,saiannadeanna2019}. The approach can be extended to learn overcomplete sparsifying transforms~\cite{ravbres13}.


\paragraph{Learning Unions of Transforms}
The union of transforms model~\cite{wensaibres15} is an extension of the single square transform case. 
Here, we assume that image patches can be grouped into multiple classes, with a different transform for each class that sparsifies the patches in that class. The model enables flexibly capturing the diversity of features in image patches with a collection or union of transforms.
The union of transforms model can be learned as follows: 
	\begin{equation} \label{eq:ultra-cost}
\begin{aligned}
\small \textrm{arg}\min_{\left \{ \omg_k, \Z, C_k \right \}}\sum_{k=1}^{K}\sum_{i\in C_k}
\left \{ \| \omg_k\mathbf{X}_i-\Z_i \|_{2}^2 + \gamma^2\|\Z_{i}\|_0 \right \} + \sum_{k=1}^{K}\lambda_k \widetilde{\mathcal{R}}(\omg_k), \text{ s.t. } \{C_k\}\in \mathcal{G},
\end{aligned}
\vspace{-0.05in}
\end{equation}
where $k$ is the cluster or group index, $K$ is the total number of clusters, $C_k$ is the set of indices of image patches grouped into the $k$th class with transform $\omg_k$, and
$\mathcal{G}$ is the set containing all possible partitions of $\{1,\cdots, S\}$ into $K$ disjoint subsets. The regularizers $\widetilde{\mathcal{R}}(\omg_k)$ $\forall$ $k$ control the properties of the learned transforms. 
Problem~\eqref{eq:ultra-cost} and its variants can be optimized by alternatingly optimizing the cluster memberships and sparse codes, and the collection of transforms, with efficiently computed solutions in each step~\cite{pwls-ultra2018,wensaibres15}.

Other variations of this model have also been explored, e.g., in~\cite{wen2017frist}, where the different transforms are related to each other by flipping and rotation operators, creating a more structured model.

\paragraph{Image Reconstruction with Learned Transforms}
We formulate the image reconstruction problem regularized by a union of pre-learned sparsifying transforms as
\begin{equation}\label{imagereconTL}
\min_{\x, \left \{ \z_j,C_k \right \}} f(\x,\y) + \beta\sum_{k=1}^{K}\sum_{j\in C_k} \big\{\|\omg_k\mathbf{P}_j\x - \z_j\|_2^2 + \gamma\|\z_j\|_0\big\}, \text{ s.t. }\{C_k\} \in \mathcal{G}.
\end{equation}
When $K=1$, this is equivalent to using a single square sparsifying transform model for reconstruction.
The algorithm for~\eqref{imagereconTL} in~\cite{pwls-ultra2018}
alternates between an image update step, i.e., updating $\x$ with the other variables fixed, and a sparse coding and clustering step, i.e., updating all the $\z_j$ and $C_k$ jointly with $\x$ fixed. 
\begin{figure}[!htp]
    \centering
    \includegraphics[width=1\linewidth]{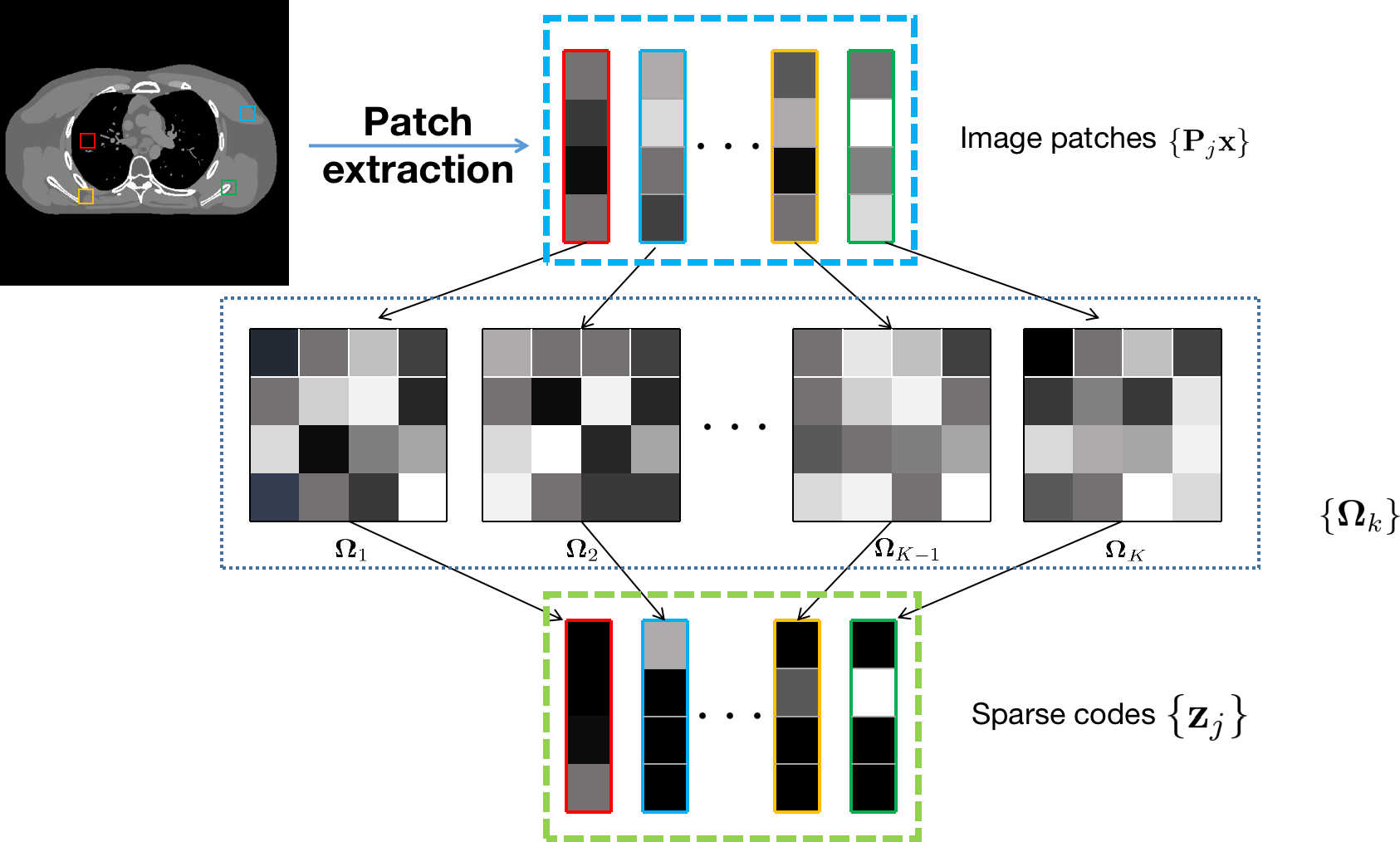}
    \caption{Illustration of ULTRA~\cite{pwls-ultra2018}. Each patch extracted from an image will be sparsified by a particular transform from the learned union of transforms that can sparsify it best.}
    \label{fig:ultra}
\end{figure}

While the sparsifying transforms can be pre-learned from images and used in~\eqref{imagereconTL} (see~\cite{pwls-ultra2018} for application in CT), they can also be learned at the time of image reconstruction (see~\cite{ravishankar:16:tci,wensailukebres19} for examples in MRI) to capture more image-adaptive features. The latter would involve optimizing more variables at reconstruction time and more computation.
A hybrid between the two approaches such as involving pre-learning a sparsifying model and then adapting it further at reconstruction time could combine the benefits of both regimes.
As a side note, while sparsifying transforms can be learned from and applied to any type of signal including image patches, for images, they can be equivalently applied directly to input images via convolutional operations~\cite{pfisbres19,wensailukebres19}.

\paragraph{Extensions: Multi-layer and Online Sparsifying Transform Learning}
The aforementioned (single-layer) sparsifying transforms can be extended to multi-layer (deep) setups to further sparsify signals or images~\cite{saibrenmulti,multi-layer-st:18,mars:yang:20}. 
Here, the transform domain residuals (obtained by subtracting the transformed data and their sparse approximations) in each layer of the multi-layer transform model are fed as input to the subsequent layer to be further sparsified.
The learning problem (using a single transform rather than a union per layer) can be formulated as
\begin{equation}
\min_{\{\omg_l,\Z_l\}} \sum_{l=1}^{L}\big\{\|\omg_l\mathbf{R}_l-\Z_l\|_F^2+\gamma_l^2\|\Z_l\|_0\big\}, \text{ s.t. }\mathbf{R}_{l+1} = \omg_{l}\mathbf{R}_{l} - \Z_{l},\ \omg_l^T\omg_l = \bm{I}, \forall l,
\end{equation} 
where $\mathbf{R}_l$ denotes the transform domain residuals in the $l$th layer (defined iteratively) for $l = 2, \cdots, L$, whereas in the first layer $\mathbf{R}_1$ denotes a matrix with the original training signals or image patches; and $\{\gamma_l>0\}$ are layer-wise parameters controlling sparsity levels. For learning simplicity, a unitary constraint is used for the transforms.
Once learned, the multi-layer model can be used to regularize image reconstruction cost function, which can outperform single-layer models~\cite{multi-layer-st:18,mars:yang:20}.
Recently, the union of transforms model was also combined with the multi-layer framework with $L=2$ layers~\cite{TwoLyr-cluster:20} for CT image reconstruction.

Finally, recent works have also studied learning sparsifying transforms and dictionaries for dynamic image reconstruction in a fully online (or time-sequential) manner~\cite{briansairajjeff18,wensaibresvidosat19}.
In the online learning framework, the learned transform (or dictionary) is evolved over time while simultaneously reconstructing a time-series of images from limited or corrupted measurements. This allows the transform to change with the data depending on the underlying dynamics, and is particularly suitable for applications such as dynamic MRI~\cite{briansairajjeff18}.
Often an optimization problem for learning and reconstruction needs to be solved every time measurements for a new frame are received. However, using ``warm starts" for the variables based on previous estimates speeds up the convergence of the algorithms (necessitating only few iterations per frame) making online algorithms both effective and efficient for dynamic imaging.

\subsection{Autoencoder}
An autoencoder usually consists of an encoding pipeline (or encoder), which extracts features from the input signal, and a decoding pipeline (or decoder), which reconstructs the signal from the extracted features. 
The basic structure of an autoencoder can be described as follows:
\begin{equation}
\begin{aligned}
&\text{Encoding: }\z = \bm{E}_{\bm{\theta}}(\x),\\
&\text{Decoding: }\widehat{\x} = \bm{D}_{\bm{\phi}}(\z),
\end{aligned}
\end{equation}
where $\x\in\mathbb{R}^n$ is an input signal or image, $\z\in\mathbb{R}^m$ represents encoded features from the encoder $\bm{E}_{\bm{\theta}}(\cdot)$ with parameters $\bm{\theta}$, $\bm{D}_{\bm{\phi}}(\cdot)$ denotes the decoder with parameters $\bm{\phi}$, and $\widehat{\x}\in\mathbb{R}^n$ is the reconstructed signal.

Autoencoders are usually trained with loss functions that describe reconstruction errors such as the mean squared error (MSE) loss or the cross-entropy loss.
A representative type of autoencoder using the reconstruction error loss is the \textit{denoising autoencoder} (DAE)~\cite{DAE:2010}.
A DAE is trained in an unsupervised manner with a set of given reference images. 
Noisy samples are generated by applying a random perturbation to the reference images,
e.g., adding Gaussian noise.
The DAE is then trained to map between noisy images and their corresponding reference images.
Additional constraints can be considered in training an autoencoder, which leads to various types of autoencoders. For example, \textit{contractive autoencoders} (CAE) constrain the encoder via adding the Frobenius norm of the Jacobian of the encoder to the basic reconstruction error training loss in order to improve the robustness of feature extraction with the encoder~\cite{CAE:11}; the \textit{sparse autoencoders} consider the sparsity constraint during training via setting the majority of the hidden units to be zeros in the encoding process~\cite{k-sparse-AE:13}.


Autoencoders have been incorporated into MBIR frameworks to provide learning-based priors, but most of these works train autoencoders in a supervised manner based on paired noisy and reference images~\cite{ravchfess17,chfess18,Chun&etal:19MICCAI,chun2019momentumnet}. MBIR with unsupervised learning-based DAE was proposed recently for MRI and sparse-view CT image reconstruction~\cite{DAE-mir-wavelet:19:liang,DAE-mir:20:liang,DAE-recon:19:liang}. Other types of unsupervised autoencoders may also have potentials to be combined with MBIR in a similar way.

\subsection{Generative Adversarial Network (GAN) based Methods}
A generative adversarial network (GAN) is composed of two models that compete with each other during training: a generator $G(\cdot)$ and a discriminator $D(\cdot)$. The generator attempts to generate realistic data, while the discriminator attempts to discriminate between generated data and real data~\cite{GAN:14:goodfellow}.
Let the inputs to a GAN be distributed as
$\z\sim p_{\z}(\z)$, where $\z$ denotes the noise variables and $p_{\z}(\z)$ is the probability distribution of $\z$. The generator maps the noise $\z$ to the data space as $G(\z)$. The generated data is expected to have a distribution close to that of the real data $\x$, i.e., $p_{\textrm{data}}(\x)$.
The discriminator outputs a single scalar that represents the probability of its input coming from real data rather than from generated data.

This idea leads to a cross entropy-based GAN optimization cost:
\begin{equation}
\min_G\max_D \mathbb{E}_{\x\sim p_{\textrm{data}}(\x)}[\log D(\x)] + \mathbb{E}_{\z\sim p_{\z}(\z)}[\log (1 - D(G(\z)))],
\end{equation}
where $D$ is trained to maximize the probability of correct labeling, i.e., maximize $D(\x)$ and $1-D(G(\z))$ (with logarithm), and $G$ is trained to minimize the term ${\log (1 - D(G(\z)))}$, all in an expected sense over the respective distributions.
Various improvements have been made to GANs in terms of loss functions and network architectures. Examples include WGAN~\cite{WGAN:17:icml}, which replaces the cross entropy-based loss with the Wasserstein distance-based loss, LSGAN~\cite{LSGAN:17} that uses the least squares measure for the loss function, the VGG19 network induced GAN~\cite{ledig2017photo}, and the cycleGAN, which incorporates the cyclic consistency loss~\cite{cycleGAN:17}.

In medical imaging, GANs have shown promising results for applications such as low-dose CT image denoising~\cite{Wolterink17,shan:18:tmi,LDCT-denoise-WGAN:18, LDCT-denoise-WGAN:20}, multiphase coronary CT image denoising~\cite{cycleGAN19}, and compressive sensing MRI~\cite{mir-recon-gan:18:tmi}.
Recently, GANs have been combined with model-based methods for image reconstruction. For example, a method for PET image reconstruction uses a pre-trained GAN to constrain an MBIR problem by encouraging the iterative reconstructions to lie in a feasible set created by outputs of the trained generator with noisy inputs. To enforce the noisy inputs to share similar noise level with training images, an additional regularizer involving the noisy inputs is introduced into MBIR, so that the inputs to the generator are optimized as well along with the iterative reconstructions~\cite{PET-recon-GAN:20}.
It is also possible to train a GAN using model-based costs. In~\cite{OT-cycleGAN-19}, an MBIR cost that embeds a generative network is used to formulate a training loss for the generator based on optimal transport~\cite{OT2008}. Particularly, the generator takes measurements as inputs and outputs images, which models the inverse path of the imaging problem. In the testing phase, images are simply reconstructed by feeding the measurements to the trained generator.


\subsection{Deep Image Prior}
Deep image prior (DIP) was proposed to implicitly capture prior information from deep networks for inverse image reconstruction problems such as denoising and super-resolution~\cite{DIP2018}.
Rather than utilizing explicit regularizers, DIP implicitly regularizes the problem via \textit{parameterization}, i.e., ${\x = f_{\bm{\theta}}(\z)}$~\cite{DIP2018}, where $f_{\bm{\theta}}(\cdot)$ represents the network with parameters $\bm{\theta}$, $\z$ is some fixed input such as random noise, and $\x$ is the unknown image to be restored.
The network in DIP is task-specific and DIP does not rely on external prior data.
Therefore, it has great potential in medical imaging tasks, which lack large amounts of training pairs. 
 
The DIP technique has been applied to reconstruct medical images, e.g., in  PET~\cite{DIP-PET:18:qzli}, dynamic MRI~\cite{DIP-MRI:19:Unser}, and CT~\cite{DIP-CT:20}. 
A general formulation for image reconstruction tasks with DIP is
\begin{equation}
\bm{\theta}^* = \textrm{arg}\min_{\bm{\theta}} \|\y - \A f_{\bm{\theta}}(\z)\|,\qquad \widehat{\x} = f_{\bm{\theta}^*}(\z), \label{dipsimple}
\end{equation}
where $\A$ and $\y$ denote the system matrix and measurements respectively as described in Section~\ref{classicalmbir}, $\z$
is the input to the network, $f_{\bm{\theta}}(\cdot)$ is the network that is updated from scratch during `training', and $\widehat{\x}$ is the final restored image.
Additional regularizers can be combined with this data-fidelity-based cost function, i.e.,
\begin{equation}
\bm{\theta}^* = \textrm{arg}\min_{\bm{\theta}} \|\y - \A f_{\bm{\theta}}(\z)\| + \beta \mathcal{R}(f_{\bm{\theta}}(\z)),\qquad \widehat{\x} = f_{\bm{\theta}^*}(\z),
\end{equation}
where $\beta$ is the regularizer parameter, and $\mathcal{R}(f_{\bm{\theta}}(\z))$ is an additional regularizer such as total variation~\cite{DIP-CT:20}.

Image reconstruction performance using DIP largely depends on the choice of network architectures~\cite{DIP2018}. 
In~\cite{DIP-PET:18:qzli}, the authors used a modified 3D U-Net~\cite{3D-U-Net:16} for PET image reconstruction; in~\cite{DIP-CT:20}, a U-Net without skip connections showed satisfying reconstruction results for CT; in~\cite{DIP-MRI:19:Unser}, the authors proposed a compound network consisting of a manifold-based network and a generative convolutional neural network for dynamic MRI reconstruction. The manifold-based network was designed to learn a mapping from a one-dimensional manifold to a more expressive latent space that can better represent temporal dependencies of the dynamic measurements.

DIP methods are quite similar to the synthesis dictionary and sparsifying transform learning-based reconstruction methods, when the sparsifying models are adapted directly based on the measurements~\cite{wensailukebres19} (also called blind compressed sensing). However, unlike the latter case, wherein iterative algorithms are designed to adequately optimize MBIR costs with convergence guarantees, DIP methods (e.g., for~\eqref{dipsimple}) may involve ad hoc early stopping of iterations to prevent overfitting.


\section{MBIR with Supervised Learning}
\label{mbirsupervised}

Deep learning methods are being increasingly deployed as alternatives to regularized image reconstruction due to their adaptiveness in various contexts. The direct-inversion end-to-end approaches, which use a neural network as an autoencoder, rely on a deep CNN to recover images from initial reconstructions (e.g., undersampled gridding reconstruction~\cite{Han_DLMRI20} or undersampled filtered back-projection reconstruction~\cite{jin:17:dcn}).

Neural networks can also be trained to replace the role of priors in classical MBIR methods. 
Instead of using manually fixed priors in the methods discussed in Section~\ref{classicalmbir}, it is tempting to tailor the prior to a given problem setting without compromising much the generalizability of such a learned prior. One representative work of constructing a learned prior by neural networks is the \textit{Neumann network}~\cite{NeumannNet_19}, an end-to-end data-driven method to solve the optimization problem~\eqref{prob::forward_model}. Motivated by expanding the matrix inverse term in the analytical solution (which is obtained from the first-order critical condition) as a Neumann series, the Neumann network has iterative blocks entailing steps of the gradient descent method. These gradient descent blocks are cascaded in the form of a residual network, and the neural network which represents the gradient of the prior is assumed to be piecewise linear. 

Besides the usage of neural networks as a direct end-to-end pipeline and a latent prior, we describe two major categories, where neural networks are combined with MBIR in a supervised manner.

\subsection{Plug-and-play}
Plug-and-play (PnP) regularization,
originally proposed in~\cite{venkatakrishnan_plug_2013},
is the use of a standard, off-the-self denoising 
algorithm as a replacement for the proximal 
operator inside an algorithm for model-based image reconstruction,
e.g., proximal gradient descent or ADMM.
Several algorithms to solve
 the optimization problem with the general form~\eqref{prob::forward_model}
 involve alternation between an operation that promotes data fidelity,
 (i.e., reduces $f(\x,\y)$)
 and an operation that accounts for the regularization
 (i.e., reduces $\mathcal{R}(\x)$).
 For example,
a proximal gradient algorithm iterates between a gradient descent step that drives the current estimate $\x^{(k)}$ to be a better fit to the observation $\y$ and a following proximal operator step that outputs a well-regulated estimate in the proximity of the resulting estimate from the gradient descent step. 
The proximal operator can be interpreted as a Gaussian denoising step in the maximum a posteriori sense. Therefore, there have been attempts to use neural networks to learn the proximal operator, essentially rendering the neural network function as a prior. 

The PnP idea has been extended in various ways,
notably with the consensus equilibrium framework~\cite{buzzard_plug_2018},
which allows the free combination of data terms
and regularizers without any reference to any variational problem at all.
The main advantage of the plug-and-play model 
is that it provides a simple way to combine 
physical system and noise models
with existing state-of-the-art \textit{denoisers}
in a principled manner with some theoretical guarantees~\cite{ryu_plug_2019}
(although weaker ones than are often offered by conventional variational approaches
based on convex optimization).
This flexibility has been used by several authors
to combine deep learning-based methods with model-based reconstruction;
we discuss several of these works here.

The work in~\cite{zhang_learning_2017}
demonstrates a straightforward approach to using a CNN as a plug-and-play regularizer.
The authors use half quadratic splitting
to arrive at an iterative reconstruction algorithm
of the form
\begin{align}  
    \vec{x}_{k+1}  &= \textrm{arg} \min_{\vec{x}} \label{eq:pnp-denoiser-1}
    \| \vec{y} - \vec{H}\vec{x} \|_2^2
    + \alpha \| \vec{x} - \vec{z}_k \|_2^2 \\
    \vec{z}_{k+1} &= \textrm{arg} \min_{\vec{z}} \alpha\|\vec{z} - \vec{x}_{k+1}\|_2^2
    + \beta \mathcal{R}(\vec{z}), \label{eq:pnp-denoiser-2}
\end{align}
where $k$ denotes iterations, $\mathcal{R}$ is an arbitrary regularization functional,
and $\alpha$ increases each iteration.
The top equation may be viewed as enforcing data consistency (because it involves the forward model, $\vec{H}$).
Following the plug-and-play methodology,
rather than solving the minimization problem to update $\vec{z}$,
the authors view the $\vec{z}$ update
as a denoising step with noise-level proportional to 
$\sqrt{\alpha/\beta}$.
They accomplish this denoising via a set of
CNNs, each trained to remove a different level of Gaussian noise from images
to account for the increasing $\alpha$ during reconstruction.
They validated their approach on
image denoising, deblurring, and super resolution,
but the same method would immediately extend to any modality
with a linear forward operator.
For example, one approach~\cite{ye_deep_2018}
tackles low-dose X-ray CT by plugging a CNN-based denoiser into an ADMM algorithm
and~\cite{zhao_multi_2020} addresses super-resolution with a similar approach.

Once one begins to plug-and-play with CNNs,
it is natural to explore  training the CNN
in conjunction with the data consistency
in a supervised manner.
There is theoretical justification for this:
even though~\eqref{eq:pnp-denoiser-2}
is a denoising problem,
the effective noise may not be Gaussian
and the formulation suggests that a MAP rather than MMSE denoiser is required~\cite{bigdeli_image_2019}.

One approach to training a CNN inside a plug-and-play structure end-to-end
is to unroll a fixed number of iterations of the plug-and-play optimization (see next subsection for detailed discussion with respect to the technique \textit{unrolling}).
If the data fidelity updated \eqref{eq:pnp-denoiser-1} can be computed in closed form \cite{qin_convolutional_2019},
or by CG \cite{aggarwal_modl_2019},
it is simple to compute gradients through the unrolled plug-and-play.
Otherwise, the data fidelity minimization \ref{eq:pnp-denoiser-1} may have to be limited to a small number of steps to permit gradient computations.
An important design decision here is whether to use the same or different CNN weights in each plug-and-play iteration;
the first is explored in~\cite{qin_convolutional_2019,aggarwal_modl_2019},
the second in \cite{admm-net2016}. 
Other works following the PnP theme include~\cite{VBW_prior13}, where the proximal step is replaced by a general denoising method such as non-local means or BM3D~\cite{kamilov_plug_2017}, 
and~\cite{Chang_projector_proximal17},  where the proximal step is designed to be a projection network which aims to project the input to the target image space and is trained in the framework of generative adversarial networks.

\subsection{Unrolling} \label{sec:unrolling}
The deep network approach is closely related to the idea of \textit{unrolling}, where the network is not necessarily a neural network. ``Unrolling" an optimization method refers to treating a series of $B$ iterates of an iterative optimization method as a single operation block to be applied to an input. In other words, a network considered in the unrolling context is a cascade of such blocks that embody the iterative steps of an optimization method (e.g., ADMM, block coordinate descent) with tunable parameters, and this cascade of unrolled blocks is then an end-to-end reconstruction pipeline to be applied to the observed images. The earliest proposed unrolled inverse problem solver is LISTA~\cite{LISTA2010}, where the authors unroll the iterative shrinkage and thresholding algorithm (ISTA)~\cite{fista2009} and aim to learn the dictionary for sparse encoding. Another work~\cite{Schmidt_ShrinkageUnrolling14} considers the image reconstruction problem~\eqref{prob::forward_model} and chooses the prior as a power function with filtered patches of the image as input. Blocks of the half-quadratic method (hq method) are cascaded as a network, and the shrinkage step in the hq method is replaced by a linear combinations of RBF kernels. To train the network, one aims to learn the filters used to convolve patches, the weight parameters for each the RBF kernels, and other parameters in the hq methods. Other typical examples of unrolling inverse problem solvers include the \textit{ADMM-Net}~\cite{admm-net2016}, which generalizes operations in classical ADMM algorithms to have learnable parameters and encloses each operation in a node and connects nodes into a data flow graph as an end-to-end network; \textit{BCD-Net}~\cite{ravchfess17,chfess18}, which involves learned parameters in unrolled block coordinate descent algorithms for reconstruction; \textit{primal-dual methods}~\cite{Oktem_primaldualUnroll18}, where proximal operators are realized by a parametrized residual neural network with the parameters learned from training; \textit{iterative reweighted least squares}~\cite{MoDl-mussels_unrolling18}, where the weight update is executed through a CNN denoiser; and \textit{approximate message passing}~\cite{AMP_unroll17}, where the denoising step in the algorithm is carried out by a trained CNN.

\section{Case Study: Combined Supervised-Unsupervised (SUPER) Learning} \label{mbircombined}
\begin{figure}[!htp]
    \centering
    \includegraphics[width=1\linewidth]{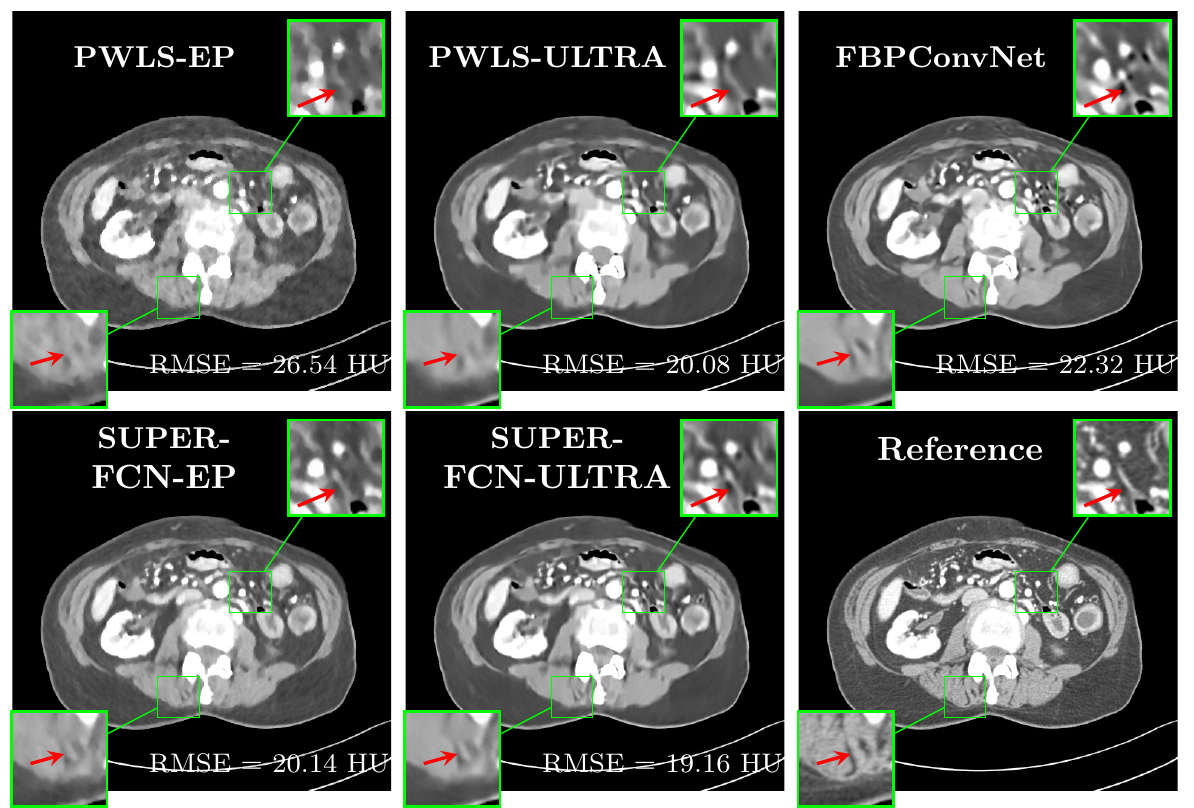}
    \caption{Low-dose X-ray CT reconstruction results with a classical MBIR method (PWLS-EP), an MBIR method with unsupervised learned prior (PWLS-ULTRA)~\cite{pwls-ultra2018}, a supervised method (FBPConvNet)~\cite{jin:17:dcn}, and their combinations using the SUPER framework (SUPER-FCN-EP and SUPER-FCN-ULTRA, where FCN refers to FBPConvNet).
	The upper right inset shows an area where unsupervised reconstruction oversmooths some details and the supervised reconstruction adds spurious features. SUPER reconstructions sharpen some details and reduce the spurious features.
	Figure adapted from~\cite{SUPER-as-submit}. The experiments are based on the NIH AAPM Mayo Clinic Low Dose CT Grand Challenge dataset~\cite{McC-Mayo}.}
    \label{fig:SUPER_results}
\end{figure}

As a case study that illustrates many of the principles in this chapter,
we take the SUPER algorithm
proposed in~\cite{li_super_2019,SUPER-as-submit}. 
SUPER combines regularization learned without supervision
along with supervised, learned regularization in the form of a CNN.
Experimental results (Figure~\ref{fig:SUPER_results}) show
that such a model can outperform both of its constituent parts.

In the following,
we will discuss how SUPER reconstruction works,
how SUPER is trained,
and present a few results comparing SUPER to
its unsupervised and supervised components
in the context of X-ray CT image reconstruction.

\subsection{SUPER Reconstruction}
SUPER reconstruction involves two main ingredients:
a variational reconstruction formulation and algorithm,
and a supervised, image-domain, learned reconstruction model.
The choice of both parts is flexible.
Here, we will focus on a combination of 
PWLS-ULTRA~\cite{pwls-ultra2018} (involving unsupervised learning) and FBPConvNet~\cite{jin:17:dcn} (learned with supervision), which we discuss in the following.

PWLS-ULTRA reconstructs images from measurements by optimizing a weighted least squares data term
combined with an ULTRA regularizer (discussed in Section~\ref{sec:sparsifying-transform})
\begin{equation}
\textrm{arg}\min_{\x}
\underbrace{
\underbrace{\|\y - \A \x\|_{\W(\y)}^2}_{f(\x, \y)}
+ 
 \underbrace{\beta \min_{\{\z_j\}, \{C_k\}} \sum_{k=1}^{K} \sum_{j\in{C_k}}
\tau_j(\|\omg_k\mat{P}_j\x-\z_j\|^2_2+\gamma^2\|\z_j\|_0)}
_{\beta\mathcal{R}(\x)}}_
{J(\x, \y)}.
\label{eq:pwls-ultra}
\end{equation}
The first term in~\eqref{eq:pwls-ultra}
is the data-fidelity term,
where 
$\| \cdot \|_{\W(\y)}$ denotes the weighted Euclidean
norm
with diagonal weight matrix
${\W(\y)} \in \mathbb{R}^{m\times m}$,
containing the estimated inverse variance of $\y$
(see~\cite{pwls-ultra2018} for further details).
The second term in~\eqref{eq:pwls-ultra}
forms a regularization term
that measures the error when the patches of $\x$ are sparse coded
by a union of transforms (see Section~\ref{sec:sparsifying-transform} for more details).
The PWLS-ULTRA algorithm reconstructs an image $\x$
from its measurements $\y$ and simultaneously clusters (in an unsupervised way) the patches of $\x$ 
by minimizing a functional of the form $f(\x, \y) + \beta \mathcal{R}(\x)$.
Note also that the sparsifying transforms, $\{\omg_k\}$,
are learned as discussed in Section~\ref{sec:sparsifying-transform}.

FBPConvNet reconstruction~\cite{jin:17:dcn}
proceeds by applying a trained CNN to an initial reconstruction as
\begin{equation} \label{eq:FBPConvNet}
    \x^* = \vec{G}_{\vec{\theta}}(\x_0(\y)),
\end{equation}
where
$\vec{G}_{\vec{\theta}}$ denotes
a CNN with parameters $\vec{\theta}$
(we discuss learning these parameters in the next section).
As originally proposed in~\cite{jin:17:dcn},
the initial reconstruction, $\x_0(\y)$,
is a pseudoinverse of $\A$,
e.g., the FBP in the case of X-ray CT.
SUPER reconstruction makes use of the fact
that this initial reconstruction is flexible;
the input to the CNN can be anything,
as long as it is the proper shape,
i.e., an element of the image (rather than measurement)
domain.

Given a variational reconstruction formulation,
e.g., the PWLS-ULTRA problem in~\eqref{eq:pwls-ultra},
and an image-domain, learned reconstruction model,
e.g., the FBPConvNet in~\eqref{eq:FBPConvNet},
SUPER combines the two in a sequential, alternating manner,
\begin{equation} \label{eq:fixed-point}
\widehat{\x}_{\bm{\theta}}^{(l)}(\y) = 
    \textrm{arg}\min_\x J(\x,\y)
    + \mu \| \x -  G_{\bm{\theta}} \left(\widehat{\x}_{\bm{\theta}}^{(l-1)}(\y) \right) \|_2^2.
\end{equation}
Equation~\eqref{eq:fixed-point}
may be understood starting from an initialization
$\widehat{\x}_{\bm{\theta}}^{(0)}(\y) = \widehat{\x}^{(0)}(\y)$,
e.g., an FBP reconstruction.
Reconstruction proceeds by applying a CNN to 
$ \widehat{\x}^{(0)}(\y)$,
then solving a variational problem to
produce $\widehat{\x}_{\bm{\theta}}^{(1)}(\y)$.
The procedure repeats for several 
iterations,
resulting in a final reconstruction.
Each iteration is called a SUPER layer.
Because the character of the 
$\widehat{\x}_{\bm{\theta}}^{(l)}(\y)$'s
change with each SUPER layer,
different network weights could be used in each
layer,
leading to the following (\textit{at testing time}) SUPER reconstruction strategy: 
\begin{equation} \label{eq:SUPER-reconstruction}
\widehat{\x}_{ \bm{\theta}^{(l)} }^{(l)}(\y) = 
     \textrm{arg}\min_\x J(\x, \y) 
    + 
    \mu \| \x - 
    G_{\bm{\theta}^{(l)}} \left(\widehat{\x}_{\bm{\theta}^{(l-1)}}^{(l-1)}(\y) \right) \|_2^2.
\end{equation}
This algorithm strategy arises from an underlying mathematical framework presented in Sections~\ref{SUPERtrain} and~\ref{SUPERinterpret}.

\subsection{SUPER Training}\label{SUPERtrain}
SUPER training involves two datasets:
a dataset of high-quality images without measurements, $\{\x_n\}_{n=1}^{N_{\text{unsup}}}$,
and a dataset of high-quality reference images
(coming from regular-dose measurements)
with corresponding
noisy (e.g., low-dose, sparse view, etc.) measurements, 
$\{\x_n, \y_n\}_{n=1}^{N_{\text{sup}}}$. 
These datasets may have different sizes, depending on the application. Complementing the models may allow exploiting limited training data.
The first dataset is used to train the unsupervised regularization term,
(e.g., ULTRA as described in the previous section);
following standard procedures from the literature (for ULTRA, see~\cite{pwls-ultra2018}).
Subsequently, the latter dataset is used to train the supervised
part of SUPER.
We focus here on the supervised training.

SUPER uses a greedy, layer-wise approach to learning weights,
solving
\begin{equation}\label{eq:super-all-sim}
\bm{\theta}^{(l)}=\textrm{arg}\min_{\bm{\theta}} 
\sum_{n=1}^{N_{\text{sup}}}
\|
G_{\bm{\theta}^{(l)}} \left(\widehat{\x}^{(l-1)}_{\bm{\theta}^{(l-1)}}(\y_n)\right) - \x_n
\|_2^2, 
\end{equation}
for $l = 1, 2, \dots, L$.
In short:
the training alternates between solving the variational problem~\eqref{eq:SUPER-reconstruction} with a fixed CNN-based reconstruction appearing in a quadratic penalty term,
and performing CNN training~\eqref{eq:super-all-sim},
with a fixed variational reconstruction as input to the CNN.
Note that the variational problem may be nonconvex or involve discrete variables (e.g., clusters) as with PWLS-ULTRA.

\subsection{Mathematical Underpinnings of SUPER} \label{SUPERinterpret}
We describe three interpretations of the above SUPER framework:
as a fixed point iteration,
as an unrolled PnP reconstruction,
and as a bilevel optimization.

To view SUPER as a fixed point iteration,
note that SUPER reconstruction~\eqref{eq:fixed-point}
is a fixed point iteration that solves the equation
\begin{equation} \label{eq:intro-SUPER}
\widehat{\x}(\y) = 
     \textrm{arg}\min_\x J(\x,\y) +
     \mu \|\x - G_{\bm{\theta}}(\widehat{\x}(\y))\|_2^2.
\end{equation}
Thus,
SUPER can be viewed as seeking a reconstruction
$\widehat{\x}$
that is the solution to a variational problem
with the solution itself involved as a regularizer.
Roughly speaking,
an $\widehat{\x}$ that solves this equation falls in between the model-based reconstruction (\eqref{eq:intro-SUPER} with $\mu=0$)
and the result of applying a CNN to the same, $G_{\bm{\theta}}(\widehat{\x}(\y))$.
While the fixed point perspective
gives a compact (if not complicated)
view of SUPER,
it cannot easily account for
changing the CNN parameters in each layer,
nor for the specifics of training.

To view SUPER as a generalized unrolled PnP reconstruction,
recall the half quadratic splitting in~\eqref{eq:pnp-denoiser-1} and~\eqref{eq:pnp-denoiser-2},
which, just like~\eqref{eq:fixed-point},
alternates between a variational problem with a quadratic penalty
and (assuming a CNN has been plugged in)
a CNN.
However, \eqref{eq:fixed-point} combines unsupervised regularizers together with the data-fidelity term and
because of the choice of the regularizer (e.g., nonconvex, nonsmooth, with discrete variables) in SUPER,
gradients may not easily propagate through the variational step.
Rather than unrolling the variational step,
SUPER follows the greedy training approach discussed in 
the previous section.
The generalized PnP perspective nicely accounts for the actual SUPER
algorithm,
but (as with all PnP regularizers)
it does not map cleanly back to a variational problem
unless the learned component happens to solve a specific denoising problem,
which is probably not the case.

Finally,
we can view the SUPER training scheme as a heuristic for solving the bilevel optimization
\begin{equation} \label{eq:bilevel}
\begin{split}
    & \textrm{arg}\min_{\bm{\theta}}
    \sum_{n=1}^{N_{\text{sup}}} \| G_{\bm{\theta}}( \widehat{\x}_{\bm{\theta}}(\y_n)) - \x_n \|_{2}^{2}
  \\
&\text{s.t.} \quad \widehat{\x}_{\bm{\theta}}(\y_n) = \textrm{arg}\min_\x J(\x, \y_n) +
     \mu \|\x - G_{\bm{\theta}}(\x)\|_2^2.
     \end{split}
\end{equation}
Specifically, training involves alternating between solving the upper-level problem with the lower-level result fixed
and then solving the lower-level problems approximately.
Roughly,
we seek CNN parameters, $\bm{\theta}$, such that when the CNN is used as regularization in a variational reconstruction,
the CNN applied to the minimizer is close to the ground truth.
Such an approach makes sense when the CNN acts as a projector onto the space of high-quality reconstruction images.
This interpretation is similar to the fixed-point one,
except that, here,
the focus is on the CNN parameters and the training process,
as opposed to a focus on the actual reconstruction.

Bilevel optimization has some links with unrolling
(Section~\ref{sec:unrolling}).
Whereas in unrolling, the \textit{$\argmin$} is approximated with a finite number of steps of a \textit{specific} minimization algorithm
with parametric weights,
some authors (in other settings) have begun to develop the tools to attack bilevel problems directly.
For example,~\cite{amos_optnet_2017} and \cite{agrawal_differentiable_2019}
provide differentiable solvers for convex optimization problems;
and \cite{peyre_learning_2011,mairal_task_2012,sprechmann_supervised_2013,chen_learning_2014}
attack specific lower-level problems.
To our knowledge,
this formulation has not been used for performing supervised biomedical image reconstruction,
thus opening a new frontier in the area.
A potential advantage is that explicitly learning components of image reconstruction formulations in a bilevel manner could then enable directly optimizing these formulations at testing time  with any suitable algorithm (until convergence),
thereby decoupling the training process from a finite unrolling of a specific algorithm.

\section{Discussion and Future Directions} \label{discussion}
This chapter focused on biomedical image reconstruction methods at the intersection of MBIR and machine learning.
After briefly reviewing classical MBIR methods for image reconstruction, we discussed the combinations of MBIR with unsupervised learning, supervised learning, or both. Such combinations offer potential advantages for learning even with limited data.

Multiple categories of unsupervised learning-based methods were reviewed, including
sparsity-based synthesis dictionary learning and sparsifying transform learning based methods (where a variety of structures can be imposed on the learned dictionaries or transforms including 
unions of models, rotation invariance, filterbank models, multi-layer models, etc.) and generic deep neural network-based methods such as based on autoencoders, or GANs, or the deep image prior approach, where networks are adapted to reconstruct specific images. Among methods at the intersection of MBIR and supervised learning, we reviewed unrolling-based methods (that learn the parameters in conventional iterative MBIR algorithms with limited iterations, in a supervised way) and CNN-based plug-and-play methods.
Finally, we discussed a recent combined supervised-unsupervised learning approach called SUPER that brings together both unsupervised priors (e.g., recent transform or dictionary learning-based) and supervised learned neural networks in a common MBIR-based framework. We interpreted this framework from the perspective of fixed point iterations, plug-and-play models, as well as a challenging bilevel learning formulation.
SUPER learning with limited training data obtains improved image reconstructions (see~\cite{SUPER-as-submit}) compared to the individual supervised or unsupervised learning-based methods by themselves, i.e., a unified approach performs better than the individual parts.

Given growing interest in research at the intersection of model-based and learning-based image reconstruction, we expect continued developement of methods and corresponding theory in this area for years to come.
First, the systematic unification of physics-based models (e.g., forward models, partial differential equations based models, etc.) and mathematical and statistical models with models learned from both simulations and experimental data is of critical importance for better leveraging data sets and domain knowledge  
in computational imaging applications. 
Second, developing provably correct algorithms to
directly and effectively learn MBIR regularizers in a bilevel optimization framework may provide an interesting counter point or alternative to unrolling-based (supervised) methods, where the choice of the algorithm to unroll and the number of iterations to unroll is currently made in a heuristic manner. 
Finally, while we focused on machine learning methods for image reconstruction, there is budding interest in machine learning for designing data acquisition as well~\cite{saibresadaptsampl13,learnCSMRI18,jacob2020:JMODL}, where the integration of model-based and learnable components could play a key role.



\bibliographystyle{abbrv}
\bibliography{egbib,refs_Mike,refs_sai}

\end{document}